\documentclass{article}
\usepackage[T1]{fontenc} 
\usepackage[utf8]{inputenc} 
\usepackage{ismir,amsmath,cite,url}
\usepackage{graphicx}
\usepackage{color}

\usepackage{lineno}

\usepackage{mathtools}
\usepackage{amsfonts, amssymb}
\usepackage{subfig}

\def\diag{\mathop{\rm diag}\nolimits}

\title{A Contextual Latent Space Model:\\ Subsequence Modulation in Melodic Sequence}

\oneauthor
 {Taketo Akama}
 {Sony Computer Science Laboratories, Tokyo, Japan\\ {\tt taketo.akama@sony.com}}

\def\authorname{T. Akama}

\usepackage[bookmarks=false,pdfauthor={\authorname},pdfsubject={\papersubject},hidelinks]{hyperref}

\begin{document}

\maketitle
\begin{abstract}
Some generative models for sequences such as music and text allow us to edit only subsequences, given surrounding context sequences, which plays an important part in steering generation interactively. However, editing subsequences mainly involves randomly resampling subsequences from a possible generation space. We propose a contextual latent space model (CLSM) in order for users to be able to explore subsequence generation with a sense of direction in the generation space, e.g., interpolation, as well as exploring variations—semantically similar possible subsequences. A context-informed prior and decoder constitute the generative model of CLSM, and a context position-informed encoder is the inference model. In experiments, we use a monophonic symbolic music dataset, demonstrating that our contextual latent space is smoother in interpolation than baselines, and the quality of generated samples is superior to baseline models. The generation examples are available online.\footnote{\url{https://contextual-latent-space-model.github.io/demo/}}
\end{abstract}
\begin{figure}[t!]
\centering
   \includegraphics[width=0.5\textwidth, trim={0.5cm 0.6cm 0cm 1.5cm}, clip]{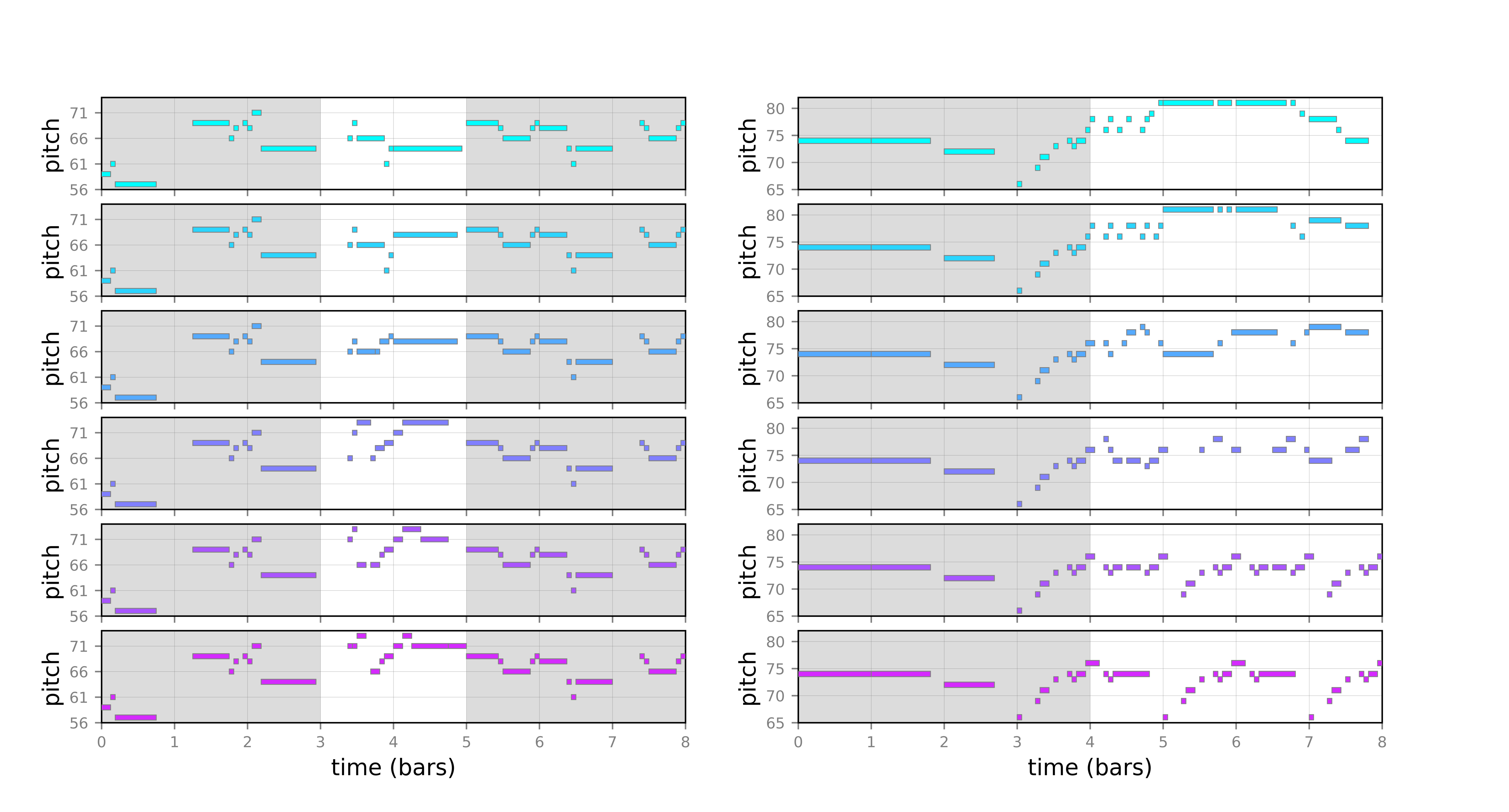}\label{fig:interp1}
  \caption{{\bf  Contextual Interpolation Examples.} Shaded regions are contexts. For both left and right figures, our CLSM first generates top and bottom melodies under constraints of contexts, and then it generates middle four interpolated points in way that is consistent with contexts. CLSM with $\beta=0.012$ is used.} 
\end{figure}
\section{Introduction}\label{sec:introduction}
Deep generative models permit sequences of decent quality to be generated such as music, lyrics, or text, where standard models generate sequences by sampling from left to right. However, to make creative works in human-machine collaborative settings, controllability—such as modifying unsatisfactory portions with specified intentions—should be improved. 

Two major model classes of controllability are (i) latent space models \cite{genseq_continuous_space,musicVAE,extres,pati2020arvae,MusicFaderNets,connectivefusion} and (ii) positional constraint models \cite{nade,BERT_Mouth,ARNN,coconet,inpaintnet}. Latent space models enable us to obtain variations or morphing/interpolations between generated sequences. Positional constraint models, on the other hand, allow us to resample a subsequence without changing the rest of the sequence ({\it context sequences}), despite the fact that subsequences are sampled randomly and cannot be controlled with morphing/interpolation or variations. Each class of models has its own benefits for making generation systems flexible. 

Can we build a hybrid model that enjoys the best of both worlds as a step towards multifaceted controllability? We propose a contextual latent space model (CLSM) that allows for positional constraints while at the same time enables latent space exploration such as interpolation or variation.
An example usage of CLSM's interpolation is narrowing down the candidates of generated subsequences given context sequences. CLSM variation can be used for obtaining minor modifications of subsequences selected among generated ones, given context sequences. 

Our approach is based on the framework of variational inference, where our CLSM is composed of prior and decoder models for the generative model and an encoder model for the inference model. The prior model of CLSM outputs a latent distribution given context sequences. We refer to the support of the distribution as the {\it contextual latent space}. The decoder model of CLSM generates subsequences that fit in with the context, given corresponding points in the contextual latent space. Finally the encoder model infers the latent space distribution given the entire sequence. 

\begin{figure*}[t!]
\centering
  \subfloat[Model Overview]{
   \includegraphics[width=0.5\textwidth, trim={0cm 0cm 0cm 0cm}, clip]{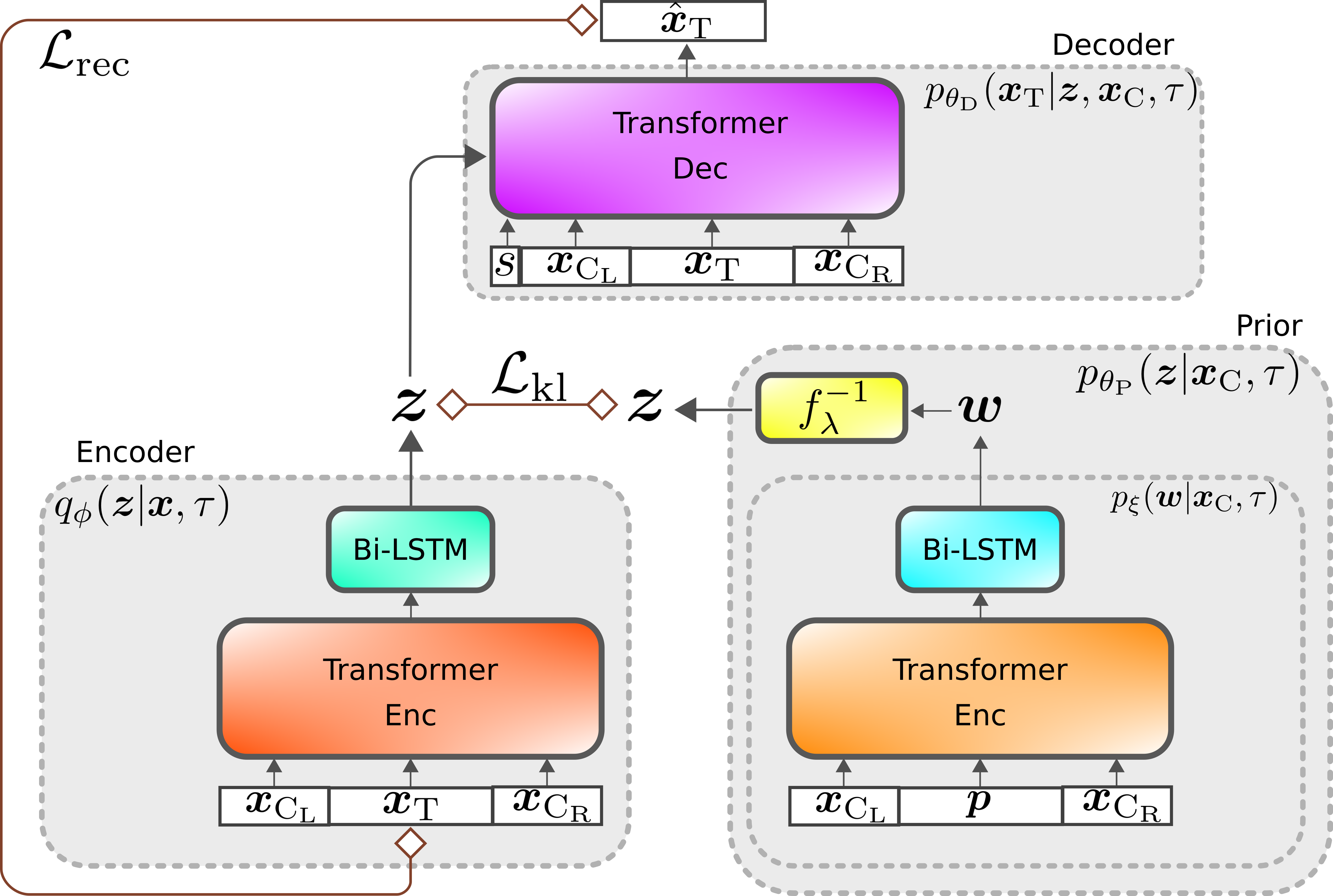}
  \label{fig:model}
  }
 \subfloat[Self-Attention Masks for Decoder Model]{
    \includegraphics[width=0.4\textwidth, trim={-10.3cm -0.3cm -10.2cm 0cm}, clip]{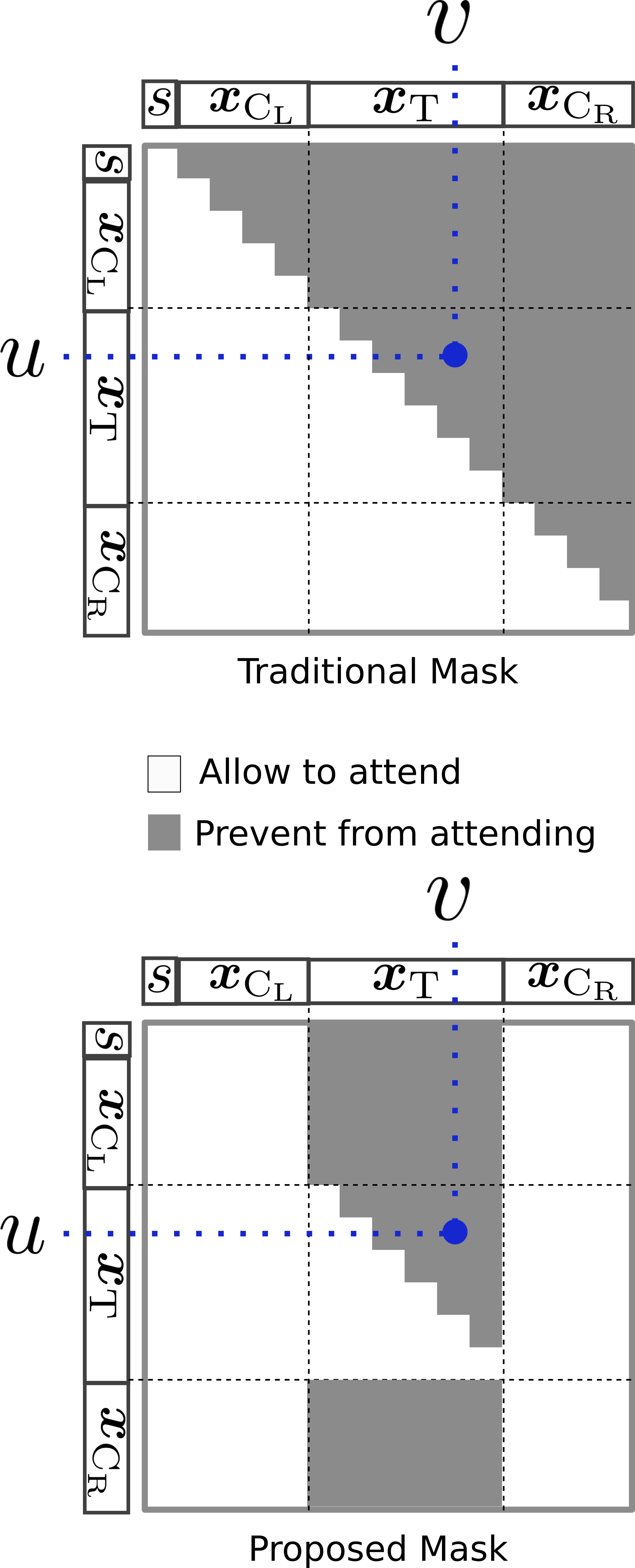}
  \label{fig:mask}
  }
  \caption{{\bf Schematic Diagram of Contextual Latent Space Model (CLSM).} (a) Linear or MLP layers are omitted for brevity. (b) Masks indicate whether position $u$ attends to position $v$.}
  \label{fig:CLSM}
\end{figure*}

We show the effectiveness of our approach using monophonic sequences in the Lakh MIDI dataset, a large symbolic music dataset \cite{lakh}. Compared with the baseline methods, our CLSM achieves better performance in terms of the smoothness of the latent space and negative log-likelihood of the generated samples. In a listening test, participants favored the generated samples of our CLSM more than those of the baselines.

Our contributions are (I) proposing a problem setting for learning a contextual latent space and providing its solution, (II) proposing novel architectures for the model, e.g., a masking strategy for the decoder and combinations of a transformer and LSTM for the encoder and prior, (III) proposing normalizing flows \cite{NormalizingFlows} for conditional priors in order to learn complex conditional priors, (IV) proposing an {\it interpolation edit distance ratio} to quantitatively assess the smoothness of latent space, and (V) demonstrating reasonable performance with our CLSM in an application for symbolic music generation.

\section{Methodology}
\subsection{Problem Scenario}\label{sec:probsen}
Let us consider an i.i.d dataset $\mathcal{D}=\{\bm{x}^{(i)}=(x^{(i)}_{1},...,x^{(i)}_{K}) \in \mathcal{A}^{K}\}_{i=1}^{N}$ of a sequence of symbols $x_{k}\in \mathcal{A}$ of length $K$, where $\mathcal{A}$ is the alphabet set of symbols. We partition each sequence $\bm{x}$ into three subsequences $\bm{x}_{\rm{C}_{\rm{L}}}$, $\bm{x}_{\rm{T}}$, and $\bm{x}_{\rm{C}_{\rm{R}}}$, such that $\bm{x}=\bm{x}_{\rm{C}_{\rm{L}}}\oplus \bm{x}_{\rm{T}}\oplus \bm{x}_{\rm{C}_{\rm{R}}}$, where $\oplus$ denotes the concatenation of sequences. We refer to subsequences $\bm{x}_{\rm{C}_{\rm{L}}}$ and $\bm{x}_{\rm{C}_{\rm{R}}}$ as {\it context sequences} and subsequence $\bm{x}_{\rm{T}}$ as the {\it target sequence}.
 Let $\tau$ denote a variable representing a set of indexes of the target sequence such that $\tau = \{|\bm{x}_{\rm{C}_{\rm{L}}}|+1,|\bm{x}_{\rm{C}_{\rm{L}}}|+2,...,|\bm{x}_{\rm{C}_{\rm{L}}}|+|\bm{x}_{\rm{T}}|\}\in \mathcal{T}$, where $\mathcal{T}$ is a set of all sets of indexes we would like to model with.
For notational simplicity, we introduce the shorthand $\bm{x}_{\rm{C}}= \{\bm{x}_{\rm{C}_{\rm{L}}},\bm{x}_{\rm{C}_{\rm{R}}}\}$.

Our goal is to train a generative model with its generative process being (1) $\tilde{\bm{z}}\sim p(\bm{z}|\bm{x}_{\rm{C}},\tau)$ and (2) $\tilde{\bm{x}}_{\rm{T}}\sim p(\bm{x}_{\rm{T}}|\tilde{\bm{z}},\bm{x}_{\rm{C}},\tau)$, where $\bm{z}\in \mathcal{Z} \subset \mathbb{R}^{d_{{\rm z}}}$ captures the variability of $\bm{x}_{\rm{T}}$ given $\bm{x}_{\rm{C}}$ and $\tau$. We would also like some distance in the latent space of the prior model to represent the similarity of $\bm{x}_{\rm{T}}$ so that the model can be used for e.g., morphing/interpolation or variation generation.

\subsection{Model}\label{sec:model}
Fig. \ref{fig:CLSM} is a schematic illustration of our model.
Our proposed approach is training a generative model by maximizing the marginal log-likelihood 
$\log p_{\theta}(\bm{x}_{\rm{T}}|\bm{x}_{\rm{C}},\tau)=\log \int p_{\theta_{\rm{D}}}(\bm{x}_{\rm{T}}|\bm{z},\bm{x}_{\rm{C}},\tau)p_{\theta_{\rm{P}}}(\bm{z}|\bm{x}_{\rm{C}},\tau)d\bm{z}$.
 Since its computation is intractable in the general case, we introduce the approximate posterior $q_{\phi}(\bm{z}|\bm{x},\tau)$ to derive the evidence lower bound (ELBO) \cite{VAE}. Formally,
\begin{align}
&\log p_{\theta}(\bm{x}_{\rm{T}}|\bm{x}_{\rm{C}},\tau)\nonumber\\
&=\log \int p_{\theta_{\rm{D}}}(\bm{x}_{\rm{T}}|\bm{z},\bm{x}_{\rm{C}},\tau)p_{\theta_{\rm{P}}}(\bm{z}|\bm{x}_{\rm{C}},\tau)d\bm{z}\nonumber\\
&=\log \int q_{\phi}(\bm{z}|\bm{x},\tau) \frac{p_{\theta_{\rm{D}}}(\bm{x}_{\rm{T}}|\bm{z},\bm{x}_{\rm{C}},\tau)p_{\theta_{\rm{P}}}(\bm{z}|\bm{x}_{\rm{C}},\tau)}{q_{\phi}(\bm{z}|\bm{x},\tau)}d\bm{z}\nonumber\\
&\geq \int q_{\phi}(\bm{z}|\bm{x},\tau) \log \frac{p_{\theta_{\rm{D}}}(\bm{x}_{\rm{T}}|\bm{z},\bm{x}_{\rm{C}},\tau)p_{\theta_{\rm{P}}}(\bm{z}|\bm{x}_{\rm{C}},\tau)}{q_{\phi}(\bm{z}|\bm{x},\tau)}d\bm{z}\nonumber\\
&= \mathcal{L}_{\rm{rec}} - \mathcal{L}_{\rm{kl}},
\end{align}
where
\begin{align}
\mathcal{L}_{\rm{rec}} &= \mathbb{E}_{q_{\phi}(\bm{z}|\bm{x},\tau)}\left[\log p_{\theta_{\rm{D}}}(\bm{x}_{\rm{T}}|\bm{z},\bm{x}_{\rm{C}},\tau)\right],\\
\mathcal{L}_{{\rm kl}} &= {\rm KL}\left(q_{\phi}(\bm{z}|\bm{x},\tau) || p_{\theta_{\rm{P}}}(\bm{z}|\bm{x}_{\rm{C}},\tau)\right).
\end{align}
In practice, we introduce weighting factors in ELBO \cite{betaVAE}. Then, our optimization problem is:
\begin{equation}
\max_{\theta, \phi} \mathbb{E}_{\bm{x}\in \mathcal{D}}\mathbb{E}_{\tau \in \mathcal{T}}\left[\frac{1}{|\bm{x}_{\rm{T}}|}\mathcal{L}_{\rm{rec}} - \beta \mathcal{L}_{\rm{kl}} \right],\label{optimization_problem}
\end{equation}
where $\frac{1}{|\bm{x}_{\rm{T}}|}$ is a normalizing factor, and $\beta$ is a balancing factor of the two terms. The specific choices of $\beta$ are explained in Sec. \ref{sec:training_settings}. 

Since the conditional prior is generally multimodal and complex, we propose modeling the conditional prior using normalizing flows \cite{NormalizingFlows}:
\begin{align}
p_{\theta_{\rm{P}}}(\bm{z}|\bm{x}_{\rm{C}},\tau)&= p_{\xi}(\bm{w}|\bm{x}_{\rm{C}},\tau)\left| \mathrm{det}\left(\frac{\partial \bm{w}}{\partial \bm{z}} \right) \right|,\label{eq:det}\\
\bm{w}&=f_{\lambda}(\bm{z})\in \mathcal{W}\subset \mathbb{R}^{d_{{\rm z}}},\label{eq:f}
\end{align}
where $f_{\lambda}\colon \mathcal{Z}\to\mathcal{W}$ is an invertible function, and $p_{\xi}(\cdot|\bm{x}_{\rm{C}},\tau)$ is a Gaussian distribution. We choose to use affine-coupling layers for $f_{\lambda}$, which was proposed for real-valued non-volume preserving (realNVP) \cite{RealNVP}. 

The Gaussian distribution and the categorical distribution are used for the encoder model $q_{\phi}(\bm{z}|\bm{x},\tau)$ and the decoder model $p_{\theta_{\rm{D}}}(\bm{x}_{\rm{T}}|\bm{z},\bm{x}_{\rm{C}},\tau)$, respectively. 

The network architectures for parametrizing each distribution are  explained in Sec. \ref{sec:architecture}.

\subsection{Applications}\label{sec:applications}
\subsubsection{Contextual Interpolation}\label{sec:interpolation}
Given $\tilde{\bm{z}}^{(1)},\tilde{\bm{z}}^{(2)}\sim p_{\theta_{\rm{P}}}(\bm{z}|\bm{x}_{\rm{C}},\tau)$,
we provide a procedure for generating contextual interpolations between $\tilde{\bm{x}}^{(1)}$ and $\tilde{\bm{x}}^{(2)}$, where $\tilde{\bm{x}}^{(i)}=\bm{x}_{\rm{C}_{\rm{L}}}\oplus \tilde{\bm{x}}_{\rm{T}}^{(i)}\oplus \bm{x}_{\rm{C}_{\rm{R}}}$ with $\tilde{\bm{x}}_{\rm{T}}^{(i)}\sim p_{\theta_{\rm{D}}}(\bm{x}_{\rm{T}}|\tilde{\bm{z}}^{(i)},\bm{x}_{\rm{C}},\tau)$ for $i=1,2$. 

First, the interpolated latent vector $\tilde{\bm{z}} (\alpha)$ with a blending ratio of $\alpha \in [0,1]$ is given by 
\begin{equation}
\tilde{\bm{z}}(\alpha)= f_{\lambda}^{-1}\left((1-\alpha)f_{\lambda}(\tilde{\bm{z}}^{(1)})+\alpha f_{\lambda}(\tilde{\bm{z}}^{(2)})\right),\label{eq:interp1}
\end{equation}
which means that linear interpolation is performed in the $\mathcal{W}$ space, but not in the $\mathcal{Z}$ space, achieving non-linear interpolation in the $\mathcal{Z}$ space.
Then, $\tilde{\bm{z}}(\alpha)$ is decoded and concatenated with the context sequence, yielding   $\tilde{\bm{x}}(\alpha)=\bm{x}_{\rm{C}_{\rm{L}}}\oplus \tilde{\bm{x}}_{\rm{T}}(\alpha)\oplus \bm{x}_{\rm{C}_{\rm{R}}}$ with 
\begin{equation}
\tilde{\bm{x}}_{\rm{T}}(\alpha)\sim p_{\theta_{\rm{D}}}(\bm{x}_{\rm{T}}|\tilde{\bm{z}}(\alpha),\bm{x}_{\rm{C}},\tau).\label{eq:interp2}
\end{equation}

\subsubsection{Contextual Variation}\label{sec:variation}
Given $\tilde{\bm{z}}\sim p_{\theta_{\rm{P}}}(\bm{z}|\bm{x}_{\rm{C}},\tau)$,
we provide a procedure for generating contextual variations of $\tilde{\bm{x}}$, where $\tilde{\bm{x}}=\bm{x}_{\rm{C}_{\rm{L}}}\oplus \tilde{\bm{x}}_{\rm{T}}\oplus \bm{x}_{\rm{C}_{\rm{R}}}$ with $\tilde{\bm{x}}_{\rm{T}}\sim p_{\theta_{\rm{D}}}(\bm{x}_{\rm{T}}|\tilde{\bm{z}},\bm{x}_{\rm{C}},\tau)$. Letting $\delta \in \mathbb{R}_{\geq 0}$ and $\bm{\epsilon} \in \mathbb{R}^{d_{\rm{z}}}$ be a scaling factor of the variation amount and sampled noise from a normal distribution $\mathcal{N}(\bm{0},\Sigma)$ with $\Sigma$ denoting the covariance matrix of $p_{\xi}(\bm{w}|\bm{x}_{\rm{C}},\tau)$,  a variation of the latent vector $\tilde{\bm{z}}$ is given by 
$\tilde{\bm{z}}(\delta)= f_{\lambda}^{-1}\left(f_{\lambda}(\tilde{\bm{z}})+\delta \bm{\epsilon}\right)$. Then, $\tilde{\bm{z}}(\delta)$ is decoded and concatenated in the same manner as Sec. \ref{sec:interpolation}.
\subsection{Network Architecture}\label{sec:architecture}

\subsubsection{Encoder Model}\label{sec:encoder}
As main architectures, we employ a two-layer ``transformer encoder'' with relative attention  \cite{relative_transformer,music_transformer}, followed by a two-layer bidirectional
LSTM network (Bi-LSTM) \cite{lstm,bi-rnn}. $\bm{x}$ is first embedded and added by a positional embedding before being inputted to the transformer. The transformer uses no masks and a sequence of vectors $\bm{E}=(\bm{e}_{1},...,\bm{e}_{K})$ is outputted. Let $\bm{E}_{\rm{T}}$ be a subsequence of $\bm{E}$, consisting of $\bm{E}$'s elements, whose indexes are in $\tau$, i.e., $\bm{E}_{\rm{T}}=(\bm{e}_{|\bm{x}_{\rm{C}_{\rm{L}}}|+1},...,\bm{e}_{|\bm{x}_{\rm{C}_{\rm{L}}}|+|\bm{x}_{\rm{T}}|})$. Only the subsequence $\bm{E}_{\rm{T}}$ is fed to the Bi-LSTM. Let $\bm{h}_{\rm{l}}$ and $\bm{h}_{\rm{r}}$ denote the last outputs of the Bi-LSTM.  They are concatenated to be fed to two multi layer perceptrons (MLPs; each for the mean and covariance of the normal distribution), yielding the encoder model $q_{\phi}(\bm{z}|\bm{x},\tau)=\mathcal{N}\left(\bm{z}|\rm{MLP}(\bm{h}_{\rm{l}}\oplus \bm{h}_{\rm{r}}),\diag(\frac{1}{2}\exp(\rm{MLP}(\bm{h}_{\rm{l}}\oplus \bm{h}_{\rm{r}})))\right)$. The MLPs consist of two layers with a SELU activation in between \cite{selu}.

Concerning the hyper-parameters for the ``transformer encoder,'' the token embedding size, the hidden size, the number of heads, and the dropout rate are set to $128$, $256$, $8$, and $0.1$, respectively. The hidden size and the dropout rate for the Bi-LSTM are set to $256$ and $0.1$, respectively. The hidden size of the MLP and the number of dimensions of $\bm{z}$ are set to $512$ and $128$, respectively.

\subsubsection{Prior Model}
As can be seen in Eq. \ref{eq:det} and Eq. \ref{eq:f}, $p_{\xi}(\bm{w}|\bm{x}_{\rm{C}},\tau)$ and $f_{\lambda}(\bm{z})$ need to be defined for the prior model.

For $p_{\xi}(\bm{w}|\bm{x}_{\rm{C}},\tau)$, as with the encoder model, a two-layer ``transformer encoder'' is followed by a Bi-LSTM. Unlike the encoder model, we replace each of the elements in  the target sequence $\bm{x}_{\rm{T}}$ with a {\it  positional constraint symbol} $p$. In other words, $\bm{x}_{\rm{C}_{\rm{L}}}\oplus \bm{p}\oplus \bm{x}_{\rm{C}_{\rm{R}}}$ is fed to the ``transformer encoder'', where $\bm{p}=(p,p,...,p)$ is a sequence of positional constraint symbols and $|\bm{p}|=|\bm{x}_{\rm{T}}|$. The hyper-parameters for the ``transformer encoder,'' the Bi-LSTM, and the number of dimensions of $\bm{z}$ are set to the same values as in Sec. \ref{sec:encoder}.

To parameterize $f_{\lambda}(\bm{z})$, four-layer affine-coupling layers are employed. Each of the scale and bias networks of the affine-coupling layers consists of a three-layer MLP, where each hidden size is $256$, and the activations are leaky ReLUs with a negative slope of $0.01$. For each of the scale networks, a $\tanh$ activation is used after the last linear layer.

\subsubsection{Decoder Model}\label{sec:decoder}
As a main architecture, we employ a two-layer ``transformer decoder'' with relative attention. Let $s$ be a symbol representing the start of a sequence. The concatenation $s\oplus \bm{x}$ is embedded and added by positional embeddings before being inputted to the transformer. 

We propose using an effective encoder-decoder attention mechanism for the latent space. Each latent vector $\tilde{\bm{z}}$ sampled from the encoder model is first fed to a linear layer to map $\tilde{\bm{z}}\in \mathbb{R}^{d_{\rm{z}}}$ to $\tilde{\bm{z}}'\in \mathbb{R}^{d_{\rm{z}} l_{\rm{z}}}$, which is reshaped to form $\tilde{\bm{Z}} \in \mathbb{R}^{d_{\rm{z}} \times l_{\rm{z}}}$, a sequence of $l_{\rm{z}}$ vectors, where the dimensionality of each vector is $d_{\rm{z}}$. The sequence $\tilde{\bm{Z}}$ is then attended to by the transformer by means of encoder-decoder attention. We set $l_{\rm{z}}=4$ in experiments.

We propose a masking strategy for modeling the decoder, as illustrated in the bottom of Fig. \ref{fig:mask}. The positions whose inputs are $s$, $\bm{x}_{\rm{C}_{\rm{L}}}$, and $\bm{x}_{\rm{C}_{\rm{R}}}$ are allowed to attend to positions except those of $\bm{x}_{\rm{T}}$. On the other hand, positions whose inputs are $\bm{x}_{\rm{T}}$ are allowed to attend to positions except the future positions of $\bm{x}_{\rm{T}}$.

Let the output of the transformer denote $\bm{D}=(\bm{d}_{1},...,\bm{d}_{K})$. Let $\bm{D}_{\rm{T}}$ be  a subsequence of $\bm{D}$, consisting of $\bm{D}$'s elements whose indexes are in $\tau$, i.e., $\bm{D}_{\rm{T}}=(\bm{d}_{|\bm{x}_{\rm{C}_{\rm{L}}}|+1},...,\bm{d}_{|\bm{x}_{\rm{C}_{\rm{L}}}|+|\bm{x}_{\rm{T}}|})$.   Let $\bm{x}_{\rm{T}}[i]$ and $\bm{D}_{\rm{T}}[i]$ denote the $i$th elements of $\bm{x}_{\rm{T}}$ and $\bm{D}_{\rm{T}}$, respectively. Then the decoder model is defined as follows: $p_{\theta_{\rm{D}}}(\bm{x}_{\rm{T}}|\bm{z},\bm{x}_{\rm{C}},\tau)=\prod_{i=1}^{|\bm{x}_{\rm{T}}|}\rm{Cat}(\bm{x}_{\rm{T}}[i]|\rm{Softmax}(\rm{Linear}(\bm{D}_{\rm{T}}[i])))$, where  $\rm{Cat}$ and $\rm{Linear}$ denote a categorical distribution and a linear layer respectively.
The hyper-parameters of the ``transformer decoder'' are set to be equal to those of the ``transformer encoder'' in Sec. \ref{sec:encoder}. 

\begin{figure*}[t!]
\centering
   \subfloat[Generation Comparison]{
    \includegraphics[width=0.40\textwidth, trim={0.6cm 0cm 1.3cm 2.5cm}, clip]{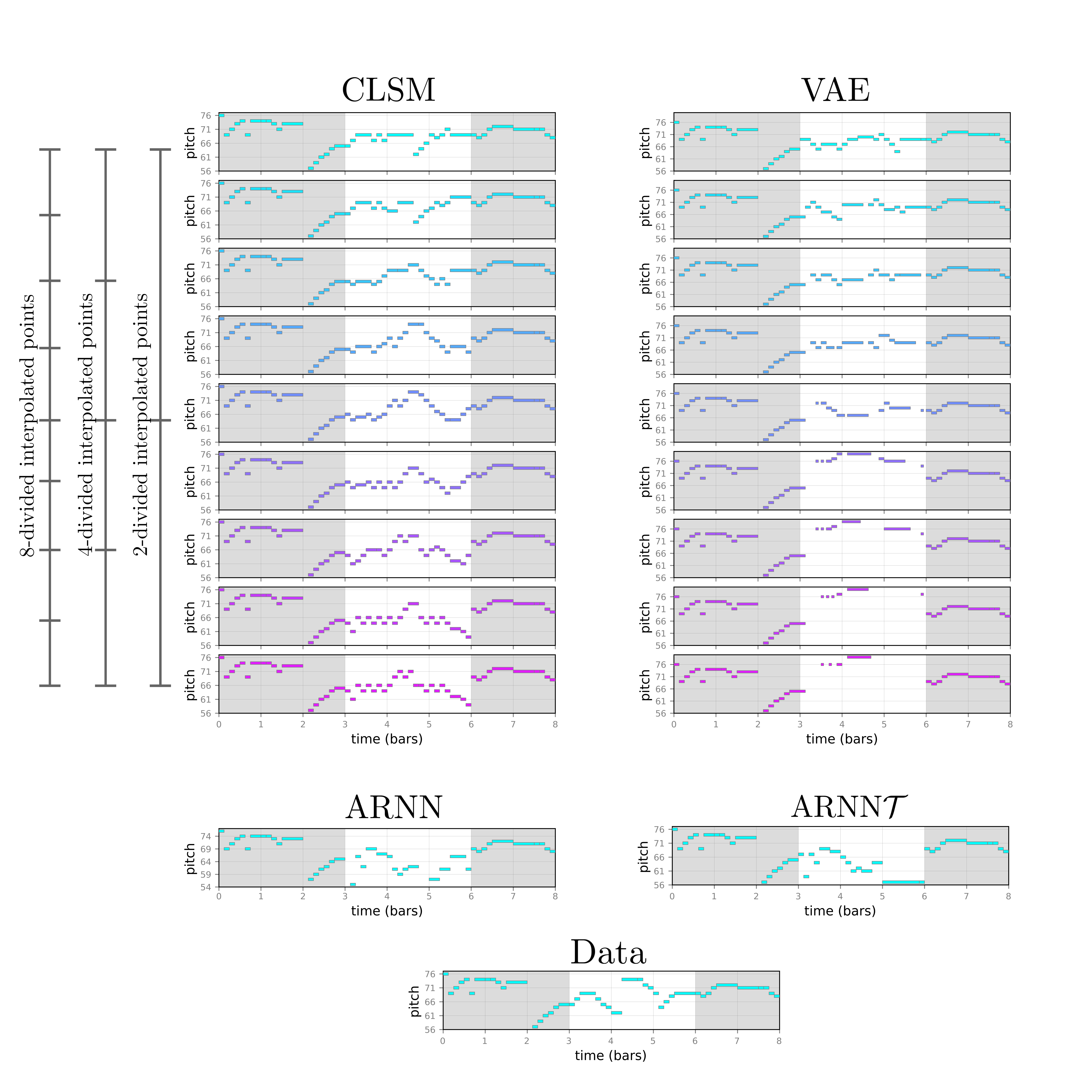}
  \label{fig:gen_comparison}
  }
  \subfloat[Objective Evaluation]{
  \includegraphics[width=0.42\textwidth, trim={0cm -3cm 0cm 1.5cm}, clip]{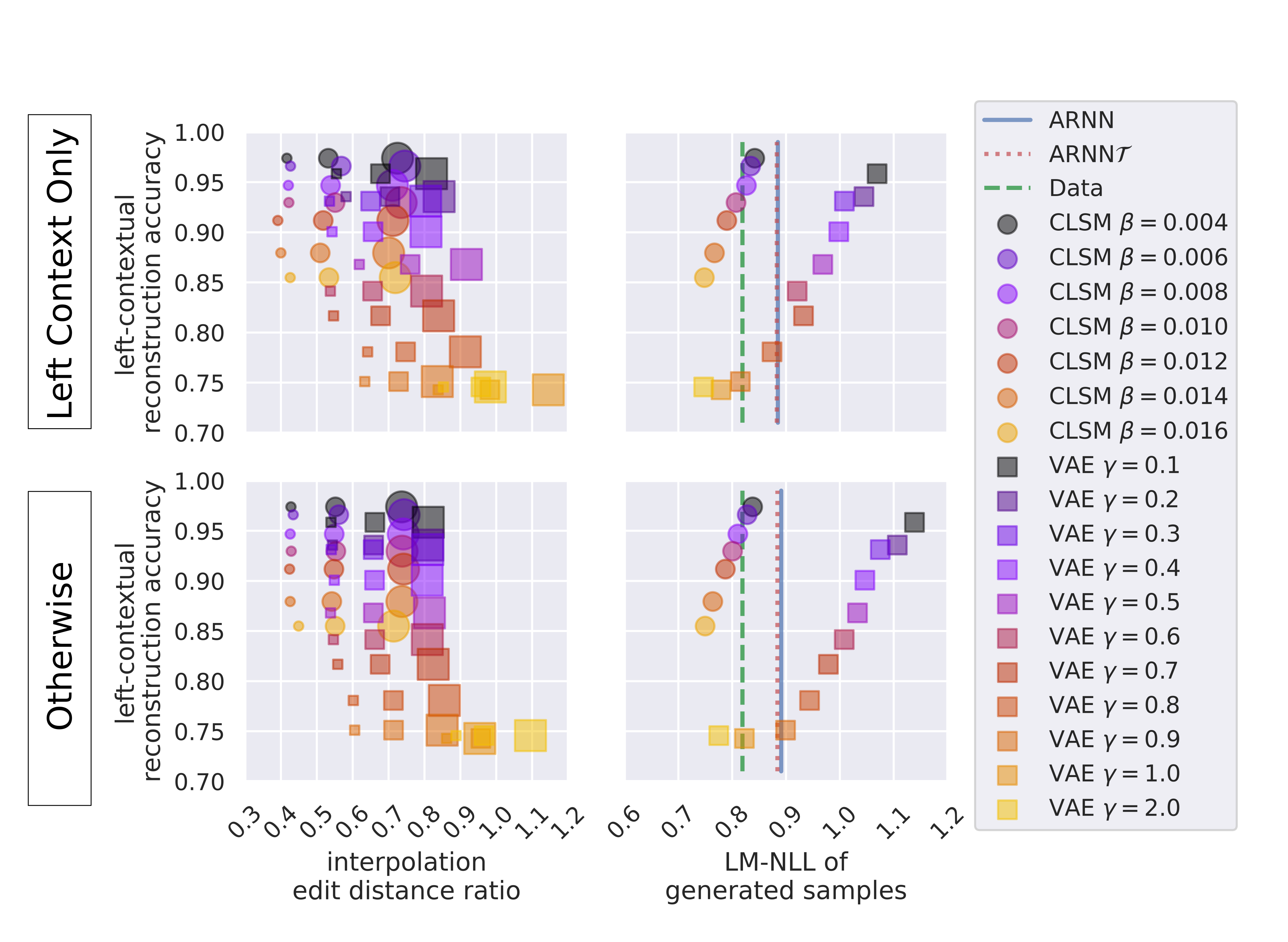}
  \label{fig:obj_eval}  }
 \subfloat[Human Evaluation]{
    \includegraphics[width=0.15\textwidth, trim={0cm -5cm 0cm 1.5cm}, clip]{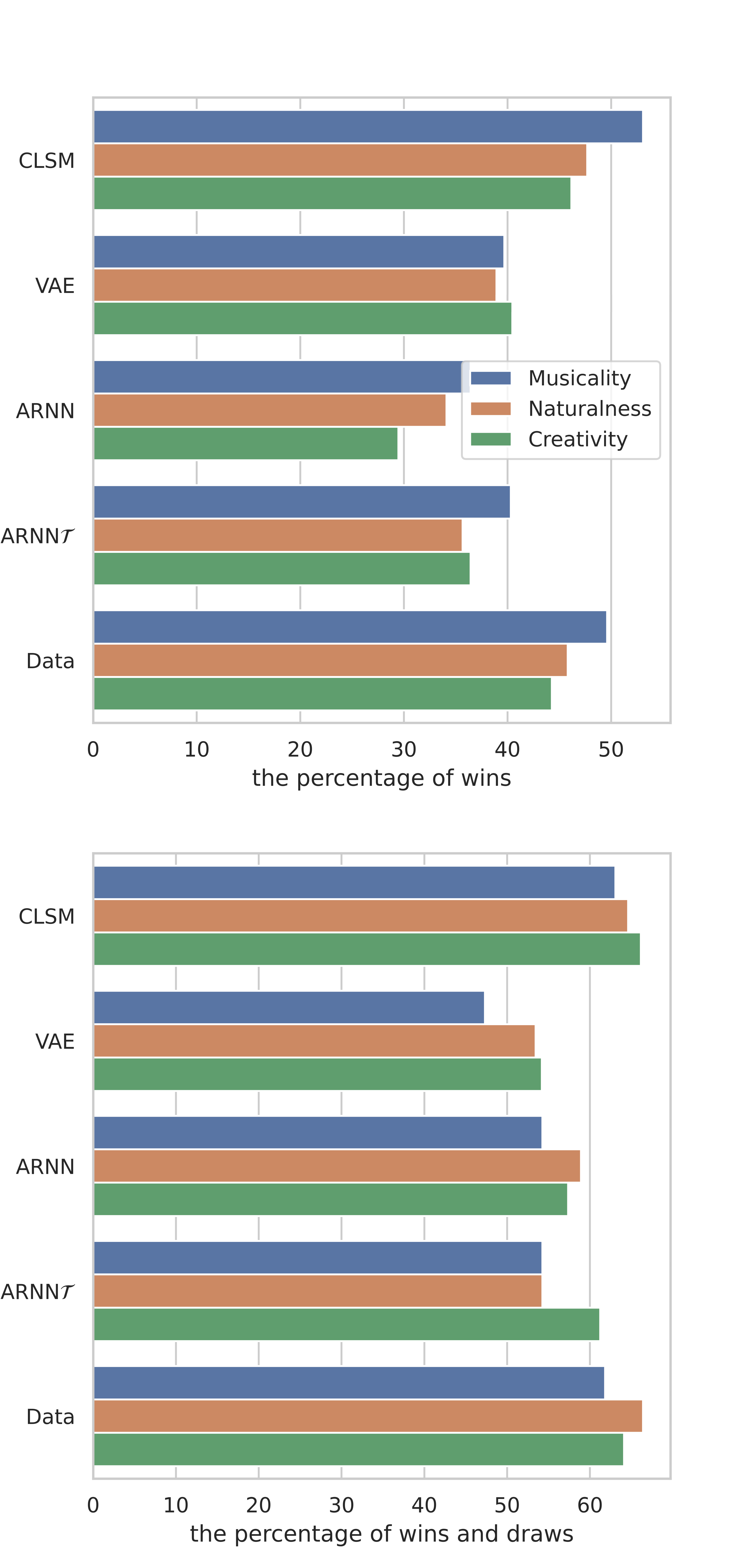}
  \label{fig:human_eval}
  }

  \caption{{\bf Generation Comparison and Experimental Results.} (a) Shaded regions are contexts.  For CLSM and VAE, $\beta=0.012$ and $\gamma=0.4$ are used. (b) For evaluating smoothness (to left), small, medium, and large marker are plots for number of divisions in interpolation $J=8,4,2$, respectively. See Sec. \ref{sec:smooth_eval} and \ref{sec:nll_eval}. (c) See Sec. \ref{sec:human_eval}.}
     \label{fig:experimental_results}
\end{figure*}

\section{Experimental Setup}\label{sec:experimental_setup}
\subsection{Dataset}\label{sec:dataset}
We created datasets from LMD-matched of the Lakh MIDI dataset \cite{lakh}, comprising 45,129 files matched to the song identity entries in the Million Song Dataset \cite{MillionSong}. Each song has one or several different versions of MIDI files. We first extracted files with a 4/4 time signature, used the accompanying tempo information to determine beat locations, and quantized each beat into 4. We then split the song identities into a proportion of 11:1:6:1:1 to create train-1, validation-1, train-2, validation-2, and test datasets, respectively. The train-1 and validation-1 datasets were for training the proposed and baseline models, whereas the train-2 and validation-2 datasets were for training evaluation models. The test dataset was for input sequences when evaluating models. We filtered out non-monophonic tracks, bass or drum tracks, and tracks outside the pitch range of [55, 84]. We  conducted data augmentation by transposing tracks to all possible keys if the transposed tracks stayed within the pitch range of [55, 84].  We retrieved 8-bar sliding windows (with a stride of 1 bar) from each track followed by filtering out windows that had more than one bar of consecutive rests. For encoding musical sequences, we adopted the melodico-rhythmic encoding proposed in \cite{ARNN}, where we used the pitch of musical notes as symbols ``55'',...,``84'' and used ``R'' to represent a rest symbol. We added an extra symbol, ``\_\_'' representing that a note is held and not replayed.

\subsection{Baseline Methods}
\subsubsection{VAE}\label{sec:vae}
We trained a VAE\cite{VAE}, in which a two-layered Bi-LSTM and LSTM were used for the encoder and decoder, respectively. The decoder, encoder, and prior models used the categorical, normal, and standard normal distribution, respectively. The optimization problem is
\begin{equation}
\max_{\psi, \omega} \mathbb{E}_{\bm{x}\in \mathcal{D}}\left[\frac{1}{|\bm{x}|}\left(\mathcal{L}_{\rm{rec}}^{\rm{VAE}}-\gamma \mathcal{L}_{\rm{kl}}^{\rm{VAE}}\right)\right],\label{optimization_problem_vae}
\end{equation}
where $\mathcal{L}_{\rm{rec}}^{\rm{VAE}}=\mathbb{E}_{q_{\omega}(\bm{z}|\bm{x})}\left[\log p_{\psi}(\bm{x}|\bm{z})\right]$, $\mathcal{L}_{\rm{kl}}^{\rm{VAE}}={\rm KL}\left(q_{\omega}(\bm{z}|\bm{x}) || p(\bm{z})\right)$, and $\frac{1}{|\bm{x}|}$ is a normalizing factor.

Given $\bm{x}_{\rm{C}}$, $\tau$, and $\tilde{\bm{z}}^{(1)},\tilde{\bm{z}}^{(2)}\sim p(\bm{z})$, interpolation is conducted as follows.
First, interpolated latent vector $\tilde{\bm{z}}(\alpha)$ is given by 
\begin{equation}
\tilde{\bm{z}}(\alpha)=(1-\alpha)\tilde{\bm{z}}^{(1)}+\alpha \tilde{\bm{z}}^{(2)}.\label{eq:interpvae1}
\end{equation}
Then, $\tilde{\bm{z}}(\alpha)$ is decoded and concatenated with the context sequence:  $\tilde{\bm{x}}(\alpha)=\bm{x}_{\rm{C}_{\rm{L}}}\oplus \tilde{\bm{x}}_{\rm{T}}(\alpha)\oplus \bm{x}_{\rm{C}_{\rm{R}}}$ with 
\begin{equation}
\tilde{\bm{x}}_{\rm{T}}(\alpha)\sim p_{\psi}(\bm{x}_{\rm{T}}|\tilde{\bm{z}}(\alpha),\bm{x}_{\rm{C}_{\rm{L}}}), \label{eq:interpvae2}
\end{equation}
where $p_{\psi}(\bm{x}_{\rm{T}}|\bm{z},\bm{x}_{\rm{C}_{\rm{L}}})$ can be immediately obtained from the autoregressive decoder model of VAE.
Note that the proposed CLSM has advantages over VAE in terms of probabilistic dependencies (i) the decoder model of VAE does not have dependencies on the right context $\bm{x}_{\rm{C}_{\rm{R}}}$, and (ii) the prior model of VAE does not have dependencies on either the left context $\bm{x}_{\rm{C}_{\rm{L}}}$ or the right context $\bm{x}_{\rm{C}_{\rm{R}}}$. The property of (i) motivates us to separately quantify the performance of models in two cases: (1) the case where only the left context exists, and (2) otherwise. In Sec. \ref{sec:smooth_eval}, Sec. \ref{sec:nll_eval}, and Fig. \ref{fig:obj_eval}, we report the performance of these two cases separately.

Random generation was conducted as follows. First, a latent vector was sampled from the prior distribution. Then, $\tilde{\bm{z}}$ was decoded and concatenated with context sequences in the same manner as the interpolation. 

The hyper-parameters of the LSTMs in VAE were set to be equal to those of the LSTMs in CLSM.

\subsubsection{ARNN}
Anticipation RNN (ARNN) \cite{ARNN} is a sequence generation model with positional constraints. Two-layered LSTMs were used for both Token-RNN and Constraint-RNN. The hyper-parameters of the LSTMs were set to be equal to those of the LSTMs in CLSM. 

\subsection{2D Plane for Comparing CLSM and VAE}\label{sec:left_con_rec}
The balancing factors $\beta$ and $\gamma$ of CLSM and VAE consist in adjusting the trade-off between the reconstruction accuracy and other model performances. Which balancing factors of CLSM correspond to which ones of VAE? It is natural to compare models of similar reconstruction accuracies or compare models of similar performances (except reconstruction accuracies). Therefore, in Sec. \ref{sec:smooth_eval}, Sec. \ref{sec:nll_eval}, and Fig. \ref{fig:obj_eval}, we plot reconstruction accuracies versus other performance metrics in 2D planes, where the more the plotted point is in the upper left corner, the better it is. As discussed in Sec. \ref{sec:vae}, CLSM and VAE have dependencies at least on the left context. 
To make the reconstruction accuracies of the CLSM and VAE mean basically identical, we used the {\it left-contextual reconstruction accuracy}, which is the average of the reconstruction accuracies of sequences where target sequences end with index $|\bm{x}|$, i.e., there is no right context and only the left context exists.

\subsection{Training Settings}\label{sec:training_settings}
For all models, teacher forcing was used, the batch size was set to $64$, and training was conducted for 2 epochs, when the losses converged. The Adam optimizer \cite{Adam} was used for all models, with the parameters $(\alpha, \beta_{1},\beta_{2})=(0.0005,0.9,0.999)$.  For CLSM or VAE, we conducted KL-annealing linearly from $\beta=0$ or $\gamma=0$ for 2 epochs \cite{genseq_continuous_space}.

\subsubsection{CLSM (Ours)}
At every iteration of the training, indexes of the target sequence $\tau = \{|\bm{x}_{\rm{C}_{\rm{L}}}|+1,|\bm{x}_{\rm{C}_{\rm{L}}}|+2,...,|\bm{x}_{\rm{C}_{\rm{L}}}|+|\bm{x}_{\rm{T}}|\} \in \mathcal{T}$ should be sampled. In the experiments, we chose to use either 1, 2, 3, or 4 bars as the target sequence length and used a stride of 1 bar for the starting index of the target sequence.  Precisely, we sampled (i) $|\bm{x}_{\rm{T}}|$ uniformly from $\{16, 32, 48, 64\}$ and then (ii) sampled $|\bm{x}_{\rm{C}_{\rm{L}}}|$ uniformly from $\{0,16, 32,...,128-|\bm{x}_{\rm{T}}|\}$. Note that $128$ corresponds to 8 bars (the sequence length) and that $16$ corresponds to $1$ bar. The balancing factor $\beta$ in Eq. \ref{optimization_problem} is set to $\{0.004,0.006,...,0.016\}$ in Secs. \ref{sec:smooth_eval}, \ref{sec:nll_eval} and $0.012$ in Sec. \ref{sec:human_eval}. The expectations of the two terms in Eq. \ref{optimization_problem} were approximated with one sample from the encoder model. 

\subsubsection{VAE (Baselines)}
The balancing factor $\gamma$ in Eq. \ref{optimization_problem_vae} is set to $\{0.1,0.2,...,1.0, 2.0\}$ in Secs. \ref{sec:smooth_eval}, \ref{sec:nll_eval} and $0.4$ in Sec. \ref{sec:human_eval}. The expectations of the reconstruction loss were approximated with one sample from the encoder model. The KL loss term was computed analytically.

\subsubsection{ARNN (Baseline)}
The vanilla ARNN is capable of imposing constraints of any positions. Since it would be possible that restricting constraints to those of our $\mathcal{T}$ would be advantageous when evaluated over $\mathcal{T}$, we also trained and evaluated this model, which we refer to as ARNN$\mathcal{T}$.

\subsection{Generation Settings}
For CLSM and VAE, each element of sequences was sampled by applying the $\rm{argmax}$ operation to the categorical distributions of the decoders. For ARNN, multinomial sampling with a temperature of $1.0$ was used.

\section{Experiments and Results}\label{sec:exp_results}

\subsection{Smoothness Analysis in Latent Space}\label{sec:smooth_eval}
To assess the smoothness of our latent space, we propose the {\it interpolation edit distance ratio} $R(J)$, which is the ratio of the distance between adjacent interpolated points (sequences) to the distance of interpolation end points (sequences). 
Formally, $R(J)$ is the normalized average edit distance $d_{\rm{edit}}(\cdot,\cdot)$ of adjacent points in $J$-divided interpolated points:
\begin{equation}
R(J)=\mathbb{E}\left[\frac{\sum^{J-1}_{j=0}d_{\rm{edit}}\left(\tilde{\bm{x}}_{\rm{T}}\left(\frac{j}{J}\right), \tilde{\bm{x}}_{\rm{T}}\left(\frac{j+1}{J}\right)\right)}{D(J-n)} \right],\label{eq:ratio}
\end{equation}
where $\tilde{\bm{x}}_{\rm{T}}(\cdot)$ is defined by Eqs. \ref{eq:interp1}, \ref{eq:interp2} for CLSM and Eqs. \ref{eq:interpvae1}, \ref{eq:interpvae2} for VAE, while $D$ is the edit distance between end points defined by $D=d_{\rm{edit}}\left(\tilde{\bm{x}}_{\rm{T}}(0), \tilde{\bm{x}}_{\rm{T}}(J)\right)$. Here, the expectation is approximated by sampling $\tilde{\bm{z}}^{(1)},\tilde{\bm{z}}^{(2)}\sim p_{\theta_{\rm{P}}}(\bm{z}|\bm{x}_{\rm{C}},\tau)$ for CLSM and $\tilde{\bm{z}}^{(1)},\tilde{\bm{z}}^{(2)}\sim p(\bm{z})$ for VAE, where 1K samples of $\bm{x}$ are uniformly sampled from the test dataset, and, for each $\bm{x}$, $\tau$ is uniformly sampled from $\mathcal{T}$. Since the edit distance $d_{\rm{edit}}\left(\tilde{\bm{x}}_{\rm{T}}\left(\frac{j}{J}\right), \tilde{\bm{x}}_{\rm{T}}\left(\frac{j+1}{J}\right)\right)$ sometimes  becomes zero, we excluded cases through division by $J-n$ instead of $J$ as in Eq. \ref{eq:ratio}, where $n$ is the number of edit distances that are zero among $j=0,1,...,J-1$. We also excluded cases where $D(J-n)=0$. The left scatter plot of Fig. \ref{fig:obj_eval} shows a comparison of CLSM and VAE. The small, medium, and large markers are plots for the number of divisions in interpolation $J=8,4,2$, respectively. For each $J$, CLSM performed better than VAE. The plots of VAE $J=8$ overlap with those of CLSM $J=4$, indicating that even $4$-divided interpolation of CLSM was as smooth as $8$-divided interpolation of VAE. The results are similar in the two cases of ``left context only'' and ``otherwise.''

\subsection{NLL Evaluation of Generated Samples}
\label{sec:nll_eval}
To evaluate the quality of generated samples, we computed the negative log-likelihood (NLL) of the generated samples. NLL was computed by using a separately trained vanilla transformer language model (LM) that had two layers and was autoregressive. The LM was trained by using the train-2 and validation-2 datasets defined in Sec. \ref{sec:dataset}. The samples to be evaluated were the same as the $8$-divided interpolated points in Sec. \ref{sec:smooth_eval}. Since ARNN is not capable of interpolation, only a random sampling of two points was performed per each context for ARNN.  The right scatter plot of Fig. \ref{fig:obj_eval} shows a comparison of the CLSM, VAE, ARNNs, and the test dataset (Data). CLSM outperformed VAE by a large margin when we compared models that were close in terms of reconstruction accuracy or NLL—a reasonable strategy of comparison as we discuss in Sec. \ref{sec:left_con_rec}. Compared with the ARNNs, the performance of CLSM was superior in all settings. The results of VAE were better in the case of ``left context only'' than in the case of ``otherwise.'' In contrast, for the CLSM and ARNNs, there were only slight differences between the two cases. This empirically demonstrates that the decoder model of VAE has dependencies only on the left context but not on the right context, as we discuss in Sec. \ref{sec:vae}.

\subsection{Human Evaluation}\label{sec:human_eval}
To further assess the quality of sequences generated by each model, we conducted listening tests using Amazon Mechanical Turk. We sampled $32$ sequences from the test dataset, for which context positions were randomly sampled, and the target sequence lengths were sampled from 1-4 bars. We conducted a pair-wise comparison of the generation results of each model using the same context sequences. We considered all possible combinations, yielding 320 pair-wise comparisons. The order within each pair was randomized. We chose to use $\beta=0.012$ for CLSM, since the performance trade-offs are well balanced according to the results in Sec. \ref{sec:smooth_eval} and Sec. \ref{sec:nll_eval}. As discussed in Sec. \ref{sec:left_con_rec} since it is reasonable to choose a VAE model with a similar reconstruction accuracy to that of CLSM, we chose $\gamma=0.4$ for VAE. Participants with different levels of musical expertise were asked to rate ``which music is better in terms of musicality, naturalness, and creativity'' on a Likert scale. Fig. \ref{fig:human_eval} shows a comparison of the CLSM, VAE, ARNNs, and the test dataset (Data). CLSM and Data performed the best in terms of the percentage of wins (Fig. \ref{fig:human_eval}, top) as well as the percentage of wins and draws (Fig. \ref{fig:human_eval}, bottom). Interestingly, the ARNNs tended to outperform VAE when the number of draws was included, whereas VAE tended to outperform the ARNNs when only the number of wins was considered. This might indicate that the performance of VAE tends to be extreme. 

\section{Related Work}\label{sec:related}
T-CVAE is a transformer-based conditional VAE model for story completion \cite{T-CVAE}. VAEAC \cite{VAEAC} is a CNN- or MLP-based VAE that enables us to impose any positional constraints. Although their probabilistic frameworks are similar to ours, the models and architectures are quite different. Unlike their models, CLSM is demonstrated to perform interpolation in the latent space. Moreover, the data domain of T-CVAE is story text, and the domains of VAEAC are image and feature classification/regression datasets, while ours is for sequence datasets and experimented on music.

Although contexts are not considered for latent variables, there are several works that use transformers for learning a global latent variable for sequences using AE or VAE. For text-style transfer, a ``transformer encoder'' outputs are all fed to GRU to yield a latent vector, which is attended to by a decoder  \cite{TransformerVAE}. To learn the styles of piano performances, a ``transformer encoder'' outputs are summed to be attended to by a decoder \cite{EncodingStyle}. In OPTIMUS for sentence modulation \cite{Optimus} and INSET \cite{INSET} for sentence infilling, a CLS token is additionally fed to a ``transformer encoder,'' and the output at the position of the CLS yields a latent vector, which is fed to a decoder either by self-attention and/or by being added to word embeddings of the decoder (OPTIMUS) or by being inputted as the first token (INSET).
 
For language processing, the authors of UniLM propose seq-to-seq LM, where they divide a whole sequence into first and second segments \cite{UniLM}. The self-attention masks are bidirectional and unidirectional for the first and second segments, respectively. Our decoder mask is different in that it divides a sequence into three segments, where the length of the first and third segments can be zero during the training or inference phase. Also, the training procedure of them is BERT-like, which is different from ours \cite{BERT}.

RealNVP has been used for the prior in VAE in order to improve the performance of VAE \cite{realNVPprior1,realNVPprior2}. However, these works are not only in the domain of images but also use non-conditional priors, which differs from ours.

\section{Conclusion}
\label{sec:conclusion}
We proposed a contextual latent space model (CLSM), in which the left and/or right contexts of sequences can be constrained to generate interpolations or variations. A context-informed prior and decoder constitute the generative model of CLSM and a context position-informed encoder is the inference model. 

The latent space of CLSM was quantitatively shown to be smoother than baselines. Furthermore, the generation fidelity was demonstrated to be superior to the baseline methods. 
It would be useful to apply our approach to other data domains such as polyphonic music, lyrics, or text.
The benefits of the latent space model are not only enabling interpolations and variations but also enabling transformations of attributes or style transfer. It would be desirable to extend our approach to these kinds of applications.

\bibliography{ismir2021}

\begin{thebibliography}{10}
\providecommand{\url}[1]{#1}
\csname url@samestyle\endcsname
\providecommand{\newblock}{\relax}
\providecommand{\bibinfo}[2]{#2}
\providecommand{\BIBentrySTDinterwordspacing}{\spaceskip=0pt\relax}
\providecommand{\BIBentryALTinterwordstretchfactor}{4}
\providecommand{\BIBentryALTinterwordspacing}{\spaceskip=\fontdimen2\font plus
\BIBentryALTinterwordstretchfactor\fontdimen3\font minus
  \fontdimen4\font\relax}
\providecommand{\BIBforeignlanguage}[2]{{%
\expandafter\ifx\csname l@#1\endcsname\relax
\typeout{** WARNING: IEEEtran.bst: No hyphenation pattern has been}%
\typeout{** loaded for the language `#1'. Using the pattern for}%
\typeout{** the default language instead.}%
\else
\language=\csname l@#1\endcsname
\fi
#2}}
\providecommand{\BIBdecl}{\relax}
\BIBdecl

\bibitem{genseq_continuous_space}
S.~R. Bowman, L.~Vilnis, O.~Vinyals, A.~Dai, R.~Jozefowicz, and S.~Bengio,
  ``Generating sentences from a continuous space,'' in \emph{Proceedings of The
  20th {SIGNLL} Conference on Computational Natural Language Learning}, 2016.

\bibitem{musicVAE}
A.~Roberts, J.~Engel, C.~Raffel, C.~Hawthorne, and D.~Eck, ``A hierarchical
  latent vector model for learning long-term structure in music,'' in
  \emph{Proc. of the 35th International Conference on Machine Learning}, 2018.

\bibitem{extres}
T.~Akama, ``Controlling symbolic music generation based on concept learning
  from domain knowledge,'' in \emph{Proceedings of the 20th International
  Society for Music Information Retrieval Conference, {ISMIR}}, 2019.

\bibitem{pati2020arvae}
A.~Pati and A.~Lerch, ``{Attribute-based Regularization of Latent Spaces for
  Variational Auto-Encoders},'' \emph{Neural Computing and Applications}, 2020.

\bibitem{MusicFaderNets}
H.~Tan and D.~Herremans, ``Music fadernets: Controllable music generation based
  on high-level features via low-level feature modelling,'' in \emph{ISMIR},
  2020.

\bibitem{connectivefusion}
T.~Akama, ``{Connective fusion: Learning transformational joining of sequences
  with application to melody creation},'' in \emph{{Proceedings of the 21st
  International Society for Music Information Retrieval Conference}}.\hskip 1em
  plus 0.5em minus 0.4em\relax ISMIR, 2020.

\bibitem{nade}
B.~Uria, I.~Murray, and H.~Larochelle, ``A deep and tractable density
  estimator,'' in \emph{Proceedings of the 31st International Conference on
  Machine Learning}, 2014.

\bibitem{BERT_Mouth}
A.~Wang and K.~Cho, ``{BERT} has a mouth, and it must speak: {BERT} as a markov
  random field language model,'' \emph{CoRR}, vol. abs/1902.04094, 2019.

\bibitem{ARNN}
G.~Hadjeres and F.~Nielsen, ``Anticipation-rnn: enforcing unary constraints in
  sequence generation, with application to interactive music generation,''
  \emph{Neural Computing and Applications}, vol.~32, no.~4, pp. 995--1005,
  2020.

\bibitem{coconet}
C.~A. Huang, T.~Cooijmans, A.~Roberts, A.~C. Courville, and D.~Eck,
  ``Counterpoint by convolution,'' in \emph{Proceedings of the 18th
  International Society for Music Information Retrieval Conference, {ISMIR}},
  2017.

\bibitem{inpaintnet}
A.~Pati, A.~Lerch, and G.~Hadjeres, ``Learning to traverse latent spaces for
  musical score inpainting,'' \emph{ISMIR}, 2019.

\bibitem{lakh}
C.~Raffel, ``Learning-based methods for comparing sequences, with applications
  to audio-to-midi alignment and matching,'' Ph.D. dissertation, 2016.

\bibitem{NormalizingFlows}
D.~Rezende and S.~Mohamed, ``Variational inference with normalizing flows,'' in
  \emph{Proceedings of the 32nd International Conference on Machine Learning},
  2015.

\bibitem{VAE}
D.~P. Kingma and M.~Welling, ``Auto-encoding variational bayes,'' \emph{CoRR},
  vol. abs/1312.6114, 2013.

\bibitem{betaVAE}
I.~Higgins, L.~Matthey, A.~Pal, C.~Burgess, X.~Glorot, M.~Botvinick,
  S.~Mohamed, and A.~Lerchner, ``beta-vae: Learning basic visual concepts with
  a constrained variational framework,'' in \emph{5th International Conference
  on Learning Representations, {ICLR}}, 2017.

\bibitem{RealNVP}
L.~Dinh, J.~Sohl{-}Dickstein, and S.~Bengio, ``Density estimation using real
  {NVP},'' in \emph{5th International Conference on Learning Representations,
  {ICLR} 2017}.

\bibitem{relative_transformer}
P.~Shaw, J.~Uszkoreit, and A.~Vaswani, ``Self-attention with relative position
  representations,'' in \emph{Proceedings of the 2018 Conference of the North
  {A}merican Chapter of the Association for Computational Linguistics: Human
  Language Technologies, Volume 2 (Short Papers)}.

\bibitem{music_transformer}
C.-Z.~A. Huang, A.~Vaswani, J.~Uszkoreit, I.~Simon, C.~Hawthorne, N.~Shazeer,
  A.~M. Dai, M.~D. Hoffman, M.~Dinculescu, and D.~Eck, ``Music transformer,''
  in \emph{International Conference on Learning Representations}, 2019.

\bibitem{lstm}
S.~Hochreiter and J.~Schmidhuber, ``Long short-term memory,'' \emph{Neural
  computation}, vol.~9, pp. 1735--80, 12 1997.

\bibitem{bi-rnn}
M.~Schuster and K.~Paliwal, ``Bidirectional recurrent neural networks,''
  \emph{IEEE Transactions on Signal Processing}, vol.~45, no.~11, pp.
  2673--2681, 1997.

\bibitem{selu}
G.~Klambauer, T.~Unterthiner, A.~Mayr, and S.~Hochreiter, ``Self-normalizing
  neural networks,'' in \emph{Advances in Neural Information Processing
  Systems}, I.~Guyon, U.~V. Luxburg, S.~Bengio, H.~Wallach, R.~Fergus,
  S.~Vishwanathan, and R.~Garnett, Eds.\hskip 1em plus 0.5em minus 0.4em\relax
  Curran Associates, Inc., 2017.

\bibitem{MillionSong}
T.~Bertin-Mahieux, D.~P. Ellis, B.~Whitman, and P.~Lamere, ``The million song
  dataset,'' in \emph{{Proceedings of the 12th International Conference on
  Music Information Retrieval ({ISMIR} 2011)}}, 2011.

\bibitem{Adam}
D.~P. Kingma and J.~Ba, ``Adam: {A} method for stochastic optimization,'' in
  \emph{3rd International Conference on Learning Representations, {ICLR}},
  2015.

\bibitem{T-CVAE}
T.~Wang and X.~Wan, ``T-cvae: Transformer-based conditioned variational
  autoencoder for story completion,'' in \emph{Proceedings of the Twenty-Eighth
  International Joint Conference on Artificial Intelligence, {IJCAI-19}}, 2019.

\bibitem{VAEAC}
O.~Ivanov, M.~Figurnov, and D.~Vetrov, ``Variational autoencoder with arbitrary
  conditioning,'' in \emph{International Conference on Learning
  Representations}, 2019.

\bibitem{TransformerVAE}
K.~Wang, H.~Hua, and X.~Wan, ``Controllable unsupervised text attribute
  transfer via editing entangled latent representation,'' in \emph{Advances in
  Neural Information Processing Systems}.\hskip 1em plus 0.5em minus
  0.4em\relax Curran Associates, Inc., 2019.

\bibitem{EncodingStyle}
K.~Choi, C.~Hawthorne, I.~Simon, M.~Dinculescu, and J.~Engel, ``Encoding
  musical style with transformer autoencoders,'' in \emph{Proceedings of the
  37th International Conference on Machine Learning}, 2020.

\bibitem{Optimus}
C.~Li, X.~Gao, Y.~Li, X.~Li, B.~Peng, Y.~Zhang, and J.~Gao, ``Optimus:
  Organizing sentences via pre-trained modeling of a latent space,'' in
  \emph{EMNLP}, 2020.

\bibitem{INSET}
Y.~Huang, Y.~Zhang, O.~Elachqar, and Y.~Cheng, ``{INSET:} sentence infilling
  with inter-sentential transformer,'' in \emph{Proceedings of the 58th Annual
  Meeting of the Association for Computational Linguistics, {ACL} 2020},
  D.~Jurafsky, J.~Chai, N.~Schluter, and J.~R. Tetreault, Eds., 2020.

\bibitem{UniLM}
L.~Dong, N.~Yang, W.~Wang, F.~Wei, X.~Liu, Y.~Wang, J.~Gao, M.~Zhou, and H.-W.
  Hon, ``Unified language model pre-training for natural language understanding
  and generation,'' in \emph{Advances in Neural Information Processing
  Systems}, vol.~32.\hskip 1em plus 0.5em minus 0.4em\relax Curran Associates,
  Inc., 2019.

\bibitem{BERT}
J.~Devlin, M.~Chang, K.~Lee, and K.~Toutanova, ``{BERT:} pre-training of deep
  bidirectional transformers for language understanding,'' in \emph{Proceedings
  of the 2019 Conference of the North American Chapter of the Association for
  Computational Linguistics: Human Language Technologies, {NAACL-HLT} 2019}.

\bibitem{realNVPprior1}
C.-W. Huang, A.~Touati, L.~Dinh, M.~Drozdzal, M.~Havaei, L.~Charlin, and
  A.~Courville, ``Learnable explicit density for continuous latent space and
  variational inference,'' 2017.

\bibitem{realNVPprior2}
H.~Xu, W.~Chen, J.~Lai, Z.~Li, Y.~Zhao, and D.~Pei, ``On the necessity and
  effectiveness of learning the prior of variational auto-encoder,'' 2019.

\end{thebibliography}

\end{document}